\pdfoutput=1

\documentclass[11pt]{article}
\usepackage[final]{sty/template/acl}
\usepackage{sty/style}

\title{Bilingual alignment transfers to multilingual alignment for unsupervised parallel text mining}
\author{
  Chih-chan Tien
  \Thanks{Work done while at the Department of Linguistics at the University of Washington.} \\
  University of Chicago \\
  \texttt{cctien@uchicago.edu} \\\And
  Shane Steinert-Threlkeld \\
  University of Washington \\
  \texttt{shanest@uw.edu} \\}

\begin{document}

\maketitle
\begin{abstract}
    This work presents methods for learning cross-lingual sentence representations using paired or unpaired bilingual texts.  We hypothesize that the cross-lingual alignment strategy is transferable, and therefore a model trained to align only two languages can encode multilingually more aligned representations. We thus introduce \emph{dual-pivot transfer}: training on one language pair and evaluating on other pairs.  To study this theory, we design unsupervised models trained on unpaired sentences and single-pair supervised models trained on bitexts, both based on the unsupervised language model XLM-R with its parameters frozen.  The experiments evaluate the models as universal sentence encoders on the task of unsupervised bitext mining on two datasets, where the unsupervised model reaches the state of the art of unsupervised retrieval, and the alternative single-pair supervised model approaches the performance of multilingually supervised models.  The results suggest that bilingual training techniques as proposed can be applied to get sentence representations with multilingual alignment.
\end{abstract}

\section{Introduction}
\label{sec:introduction}

Cross-lingual alignment as evaluated by retrieval tasks has been shown to be present in the representations of recent massive multilingual models
which are not trained on bitexts (\cite{pires2019multilingual, conneau2020emerging}).
Other studies further show that sentence representations with higher cross-lingual comparability can be achieved
by training a cross-lingual mapping (\cite{aldarmaki2019contextaware}) or fine-tuning (\cite{cao2020multilingual}) for every pair of languages.
These two lines of research show that,
on the one hand, multilingual alignment arises from training using monolingual corpora alone,
and, on the other, bilingual alignment can be enhanced by training on bitexts of specific language pairs.

Combining these insights yields a question: can training with bilingual corpora help improve \emph{multilingual} alignment?
Given a language model encoding texts in different languages with some shared structure already,
we can expect that the model further trained to align a pair of languages
will take advantage of the shared structure
and will therefore generalize the alignment strategy to other language pairs.
From a practical point of view, bitexts for some pairs of languages are more abundant than others,
and it is therefore efficient to leverage data from resource-rich pairs for the alignment of resource-poor pairs
in training multilingual language models.

To better understand the cross-lingual structure from the unsupervised models,
we also ask the following question:
how can multilingual alignment information be extracted from the unsupervised language models?
Unsupervised multilingual models out-of-the-box as sentence encoders fall short of their supervised counterparts such as LASER (\cite{artetxe2019massively})
in the task of bitext mining (\cite{hu2020xtreme}).
The discovery of cross-lingual structure in the hidden states in the unsupervised model (\cite{pires2019multilingual, conneau2020emerging}), however,
raises the possibility that with relatively light post-training for better extraction of deep features,
the unsupervised models can generate much more multilingually aligned representations.

In this paper, we address both questions with the design of \emph{dual-pivot transfer},
where a model is trained for bilingual alignment but tested for multilingual alignment.
And we hypothesize that training to encourage similarity between sentence representations from two languages,
the {dual pivots},
can help generate more aligned representations not only for the pivot pair,
but also for other pairs.

In particular,
we design and study a simple extraction module
on top of the pretrained multilingual language model XLM-R (\cite{conneau2020unsupervised}).
To harness different training signals,
we propose two training architectures.
In the case of training on unpaired sentences,
the model is encouraged by adversarial training to encode sentences from the two languages with similar distributions.
In the other case where bitexts of the two pivot languages are used,
the model is encouraged to encode encountered parallel sentences similarly.
Both models are then transferred to language pairs other than the dual pivots.
This enables our model to be used for unsupervised bitext mining,
or bitext mining where the model is trained only on parallel sentences from a single language pair.

The experiments show that both training strategies are effective,
where the unsupervised model reaches the state of the art on completely unsupervised bitext mining,
and the one-pair supervised model approaching the state-of-the-art multilingually-supervised language models in one bitext mining task.

Our contributions are fourfold:
\begin{itemize}
    \item This study proposes effective methods of bilingual training using paired or unpaired sentences
          for sentence representation with multilingual alignment.
          The strategies can be incorporated in language model training for greater efficiency in the future.
    \item The work demonstrates that the alignment information in unsupervised multilingual language models is extractable
          by simple bilingual training of a light extraction module (without fine-tuning)
          with performance comparable to fully supervised models and reaching the state of the art of unsupervised models.
    \item The models are tested using a new experimental design---\emph{dual-pivot transfer}---to evaluate the generalizability of a bilingually-supervised sentence encoder to the task of text mining for other language pairs on which it is not trained.
    \item This study shows that unsupervised bitext mining has strong performance which is comparable to bitext mining by a fully supervised model,
          so the proposed techniques can be applied to augment bilingual corpora for data-scarce language pairs in the future.
\end{itemize}

\section{Related work}
\label{sec:related work}

\paragraph{Alignment with adversarial nets}
This work follows the line of previous studies which use adversarial networks (\abbrv{gan}s) (\cite{goodfellow2014generative})
to align cross-domain distributions of embeddings without supervision of paired samples,
in some cases in tandem with cycle consistency (\cite{zhu2017unpaired}),
which encourages representations \enquote{translated} to another language then \enquote{translated} back
to be similar to the starting representations.
\textcite{conneau2018word}'s MUSE project trains a linear map from the word-embedding space of one language to that of another using \abbrv{gan}s,
the method of which is later applied to an unsupervised machine translation model (\cite{lample2018unsupervised}).
Cycle consistency in complement to adversarial training has been shown to be effective
in helping to learn cross-lingual lexicon induction (\cite{zhang2017adversarial, xu2018unsupervised, mohiuddin2020unsupervised}).
Our work is the first to our knowledge to apply such strategy of adversarial training and cycle consistency to the task of bitext mining.

\paragraph{Alignment with pretrained LMs}
We adopt the training strategy aforementioned on top of pretrained multilingual language models,
the extractability of multilingual information from which has been studied in several ways.
\textcite{pires2019multilingual} find multilingual alignment
in the multilingual BERT (mBERT) model (\cite{devlin2019bert}) pretrained on monolingual corpora only,
while \textcite{conneau2020emerging} identify shared multilingual structure in monolingual BERT models.
Other work studies the pretrained models dynamically by
either fine-tuning the pretrained model for cross-lingual alignment (\cite{cao2020multilingual})
or learning cross-lingual transformation (\cite{aldarmaki2019contextaware})
with supervision from aligned texts.
Recently, \textcite{yang2020multilingual} use multitask training to train multilingual encoders
focusing on the performance on retrieval,
and \textcite{reimers2020making} use bitexts to tune multilingual language models
and to distill knowledge from a teacher model which has been tuned on paraphrase pairs.
Also, \textcite{chi2021infoxlm} pretrain an alternative XLM-R on a cross-lingual contrastive objective.
Our work falls in the line of exploring multilinguality of pretrained models
with a distinct emphasis on investigating the multilingual structure induced by bilingual training without fine-tuning or alternative pretraining.

\paragraph{Unsupervised parallel sentence mining}
\label{paragraph:related work unsupervised parallel sentence mining}
The evaluation task of our work is bitext mining without supervision
from any bitexts or from bitexts of the pair of languages of the mining task.
Such experiments have been explored previously.
\textcite{hangya2018unsupervised} show that unsupervised bilingual word embeddings are effective on bitext mining,
and \textcite{hangya2019unsupervised} further improve the system with a word-alignment algorithm.
\textcite{kiros2020contextual} trains a lensing module over mBERT for the task of natural language inference (\abbrv{nli})
and transfers the model to bitext mining.
\textcite{keung2020unsupervised}'s system uses bootstrapped bitexts to fine-tune mBERT,
while \textcite{kvapilikova2020unsupervised}'s system uses synthetic bitexts from an unsupervised machine translation system
to fine-tune XLM (\cite{lample2019crosslingual}).
Results from the three aforementioned studies are included in Section~\ref{sec:evaluations} for comparisons.
Methodologically,
our approach differs from the above in that our system is based on another pretrained model XLM-R (\cite{conneau2020unsupervised}) without fine-tuning,
for one of the goals of the study is to understand the extractability of the alignment information from the pretrained model;
and our model receives training signals from existing monolingual corpora or bitexts,
instead of from \abbrv{nli}, bootstrapped, or synthesized data.

\begin{figure}
    \centering
    \includegraphics[scale=0.7]{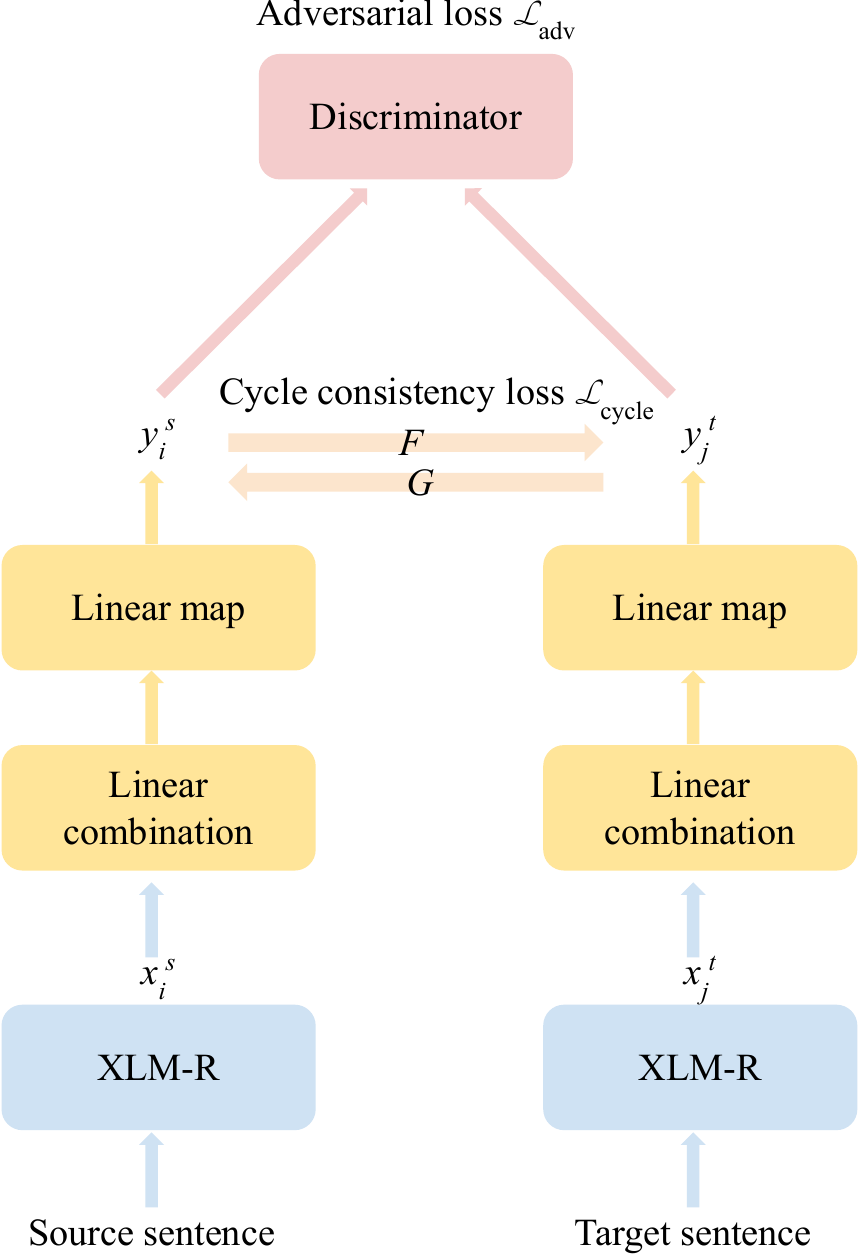}
    \caption{\label{fig:model}
        Schematic depiction of the unsupervised model with adversarial and cycle consistency losses.
    }
\end{figure}

\section{Model}

\subsection{A linear combination and a linear map}

The model as an encoder generates fixed-length vectors as sentence representations
from the the hidden states of the pretrained multilingual language model XLM-R (\cite{conneau2020unsupervised}).
Formally, given a sentence \(\pi_{i, \gamma}\) in language \(\gamma \in \lbrace s, t \rbrace\),
with the pretrained language model producing features \(x_i^\gamma\) of \(l\) layers, sequence length \(q\), and embedding size \(d\),
the extraction module \(f(\cdot)\) generates a sentence embedding \(y_i^\gamma\) of fixed size \(d\) based on the features \(x_i^\gamma\),
or
\[
    f(x_i^\gamma) = y_i^\gamma , \quad x_i^\gamma \in \mathbb{R}^{l \times q \times d} \text{ and } y_i^\gamma \in \mathbb{R}^{d}
    \, \text{.}
\]

With the parameters of XLM-R frozen, within the extraction module \(f(\cdot)\) are only two trainable components.
The first is an ELMo-style trainable softmax-normalized weighted linear combination module (\cite{peters2018deep}),
and the second being a trainable linear map.
The linear combination module learns to weight the hidden states of every layer \(l\) of the pretrained language model and output a weighted average,
on which a sum-pooling layer is then applied to \(q\) embeddings.
And then the linear map takes this bag-of-word representation and produces the final sentence representation \(y_i^\gamma\) of the model.

\subsection{Adversarial learning with unpaired texts}

Monolingual corpora in different languages with language labels provide the signal for alignment
if the semantic contents of the utterances share similar distributions across corpora.
In order to exploit this information,
we introduce to the model adversarial networks (\cite{goodfellow2014generative})
with cycle consistency (\cite{zhu2017unpaired}),
to promote similarity in the distribution of representations.

As is usual in \abbrv{gan}s, there is a discriminator module \(d(\cdot)\),
which in this model consumes the representation \(y_i^\gamma\) and outputs continuous scores for the language identity \(\gamma\) of the sentence \(\pi_{i, \gamma}\).
Following \textcite{romanov2019adversarial}, as inspired by Wasserstein-\abbrv{gan} (\cite{arjovsky2017wasserstein}),
the loss of the discriminator \(\mathcal{L}_\ind{disc}\) is the difference between the unnormalized scores instead of the usual cross-entropy loss, or
\[
    \mathcal{L}_\ind{disc} = d(y_i^{s}) - d(y_j^{t})
    \, \text{.}
\]
And the adversarial loss
\(\mathcal{L}_\ind{adv} = - \mathcal{L}_\ind{disc}\)
updates the parameters of the extraction module \(f(\cdot)\)
to encourage it to generate encodings abstract from language-specific information.

Adversarial training helps learning aligned encodings across languages at the distributional level.
At the individual level, however,
the model is not constrained to generate encodings which are both aligned and discriminative (\cite{zhu2017unpaired}).
In particular, a degenerate encoder can produce pure noise which is distributively identical across languages.
A cycle consistency module inspired by \textcite{zhu2017unpaired} is therefore used to constrain the model to encode with individual-discerning alignment.
Cycle consistency is also reminiscent of the technique of using back-translation for unsupervised translation systems (\cite{lample2018phrase}).

In this model,
a trainable linear map \(F(\cdot)\) maps elements from the encoding space of one language to the space of the other,
and another linear map \(G(\cdot)\) operates in the reverse direction.
The cycle loss so defined is used to update parameters for both of the cycle mappings and the encoder:
\[
    \mathcal{L}_\ind{cycle}
    = h(y_i^{s}, G(F(y_i^{s})))
    + h(F(G(y_j^{t})), y_j^{t})
    \, 
\]
where \(h\) is the triplet ranking loss function which sums the hinge costs in both directions:
\[
    \begin{aligned}
        h(a, b) =
        \sum_{n}
          & \max(0, \alpha -  \func{sim}(a, b) + \func{sim}(a_{n}, b)) \\
        + & \max(0, \alpha -  \func{sim}(a, b) + \func{sim}(a, b_{n}))
        \, \text{,}
    \end{aligned}
\]
where the margin \(\alpha\) and the number of negative samples \(n\) are hyperparameters,
and \(\func{sim}(\cdot)\) is cosine similarity.
The loss function \(h\) encourages the model to encode
similar representations between positive pairs \((a, b)\)
and dissimilar representations between negative pairs \((a_{n}, b)\) and \((a, b_{n})\),
where \(a_{n}\) and \(b_{n}\) are sampled from the embeddings in the mini-batch.
Based on the findings that the hard negatives,
or non-translation pairs of high similarity between them,
are more effective than the sum of negatives in the ranking loss (\cite{faghri2018vse}),
our system always includes in the summands
the costs from the hardest negatives in the mini-batch
along with the costs from any other randomly sampled ones.

The full loss of the unsupervised model is
\[
    \mathcal{L}_\ind{unsup} = \mathcal{L}_\ind{adv} + \lambda \mathcal{L}_\ind{cycle}
    \, \text{,}
\]
with a hyperparameter \(\lambda\).
This unsupervised model is presented schematically with Figure~\ref{fig:model}.

\subsection{Learning alignment with bitexts}

In addition to the completely unsupervised model,
we also experiment with a model which is supervised with bitext from one pair of languages and then transferred to other pairs.
In this set-up, instead of using cyclical mappings,
bitexts provide the alignment signal through the ranking loss directly,
so the loss for the supervised model is
\[
    \mathcal{L}_\ind{sup} = h(y_i^{s}, y_i^{t})
    \, \text{,}
\]
where \(y_i^{s}\) and \(y_i^{t}\) are representations of parallel sentences.

\section{Training}

\begin{table}
    \centering
    { \small
        \begin{tabular}{@{}lll@{}}
            \toprule
            Hyperparameter           &             & values              \\ \midrule
            output dimension         & \(d\)       & \(\{1024\}\)        \\
            \(\#\) negative samples  & \(n\)       & \(\{1, 2, 4\}\)     \\
            margin value             & \(\alpha\)  & \(\{0, 0.2, 0.4\}\) \\
            weight for cycle loss    & \(\lambda\) & \(\{1, 5, 10\}\)    \\
            discriminator step times & \(\kappa\)  & \(\{1, 2\}\)        \\ \bottomrule
        \end{tabular}
    }
    \caption{\label{tab:hyperparameters}
        Hyperparameters and experimented values.
    }
\end{table}

The model is trained with the Adam optimizer (\cite{kingma2014adam}) and learning rate \(0.001\)
with the parameters in XLM-R frozen.
Our training program is built upon
AllenNLP (\cite{gardner2018allennlp}), HuggingFace Transformers (\cite{wolf2020huggingfaces}), and PyTorch (\cite{paszke2019pytorch}).
The code for this study is released publicly.\footnote{The repository at \url{https://github.com/cctien/bimultialign}.}

For adversarial training,
the discriminator is updated \(\kappa\) times for every step of backpropagation to the encoder.
Other hyperparameters include
the dimension of the output representations \(d\),
number of negative samples \(n\),
margin value \(\alpha\),
and weight of the cycle loss \(\lambda\).
The hyperparameters and the the values which are experimented with are summarized in Table~\ref{tab:hyperparameters}.
We empirically determine the hyperparameters among experimented values,
and report their values in specific evaluation sections.

The bilingual corpora used to train the encoder is taken from OPUS (\cite{tiedemann2012parallel})
as produced for the training of XLM (\cite{lample2019crosslingual}).\footnote{We use the script \url{https://github.com/facebookresearch/XLM/blob/main/get-data-para.sh} to get the corpora MultiUN (\cite{eisele2010multiun}) and EUbookshop, where each training corpus we use is of 9 million sentences. }
We experimented with two language pairs for training the model---Arabic-English (ar-en) and German-English (de-en)---to explore potential effects of the choice of the dual-pivots.
After being trained, the encoder is evaluated on two tasks of bitext mining between texts in English and in another language.
Additionally, we train the models with the pivot pair of Arabic-German (ar-de), which does not include English,
to be evaluated on the second task.

\section{Evaluations}
\label{sec:evaluations}

\begin{table}
    \centering
    { \small
        \setlength{\tabcolsep}{2pt} 
        \begin{tabular}{@{}llllll@{}}
            \toprule
            Model                                  &                    & \multicolumn{4}{c}{F1 score (\(\%\))\hspace*{1em} xx\(\leftrightarrow\)en}                                                                \\
            \cmidrule{3-6}
                                                   & Average            & de                                                                         & fr                 & ru                 & zh                 \\
            \midrule
            \emph{Unsupervised}                    &                    &                                                                            &                    &                    &                    \\
            Model (ar-en unsup.)                   & \(81.8\)           & \(84.1\)                                                                   & \(77.9\)           & \(\bestgrp{87.9}\) & \(\bestgrp{77.1}\) \\
            Model (de-en unsup.)                   & \(\bestgrp{82.4}\) & \(\bestgrp{91.4}\)                                                         & \(75.6\)           & \(86.1\)           & \(76.5\)           \\
                                                   &                    &                                                                            &                    &                                         \\
            XLM-R L16-boe                          & \(68.7\)           & \(75.4\)                                                                   & \(65.0\)           & \(75.6\)           & \(59.0\)           \\
            \textcite{kvapilikova2020unsupervised} & \(75.8\)           & \(80.1\)                                                                   & \(\bestgrp{78.8}\) & \(77.2\)           & \(67.0\)           \\
            \textcite{keung2020unsupervised}       & \(69.5\)           & \(74.9\)                                                                   & \(73.0\)           & \(69.9\)           & \(60.1\)           \\
            \textcite{kiros2020contextual}         & \(51.7\)           & \(59.0\)                                                                   & \(59.5\)           & \(47.1\)           & \(41.1\)           \\
            \midrule
            \emph{One-pair supervised}             &                    &                                                                            &                    &                    &                    \\
            Model (ar-en bitexts sup.)             & \(89.1\)           & \(91.7\)                                                                   & \(89.2\)           & \(90.1\)           & \(85.6\)           \\
            Model (de-en bitexts sup.)             & \(\bestgrp{89.6}\) & \(\bestgrp{92.5}\)                                                         & \(\bestgrp{89.6}\) & \(\bestgrp{90.3}\) & \(\bestgrp{85.8}\) \\
            \midrule
            \emph{Fully Supervised}                &                    &                                                                            &                    &                    &                    \\
            LASER                                  & \(92.8\)           & \(95.4\)                                                                   & \(92.4\)           & \(92.3\)           & \(91.2\)           \\
            LaBSE                                  & \(\bestgrp{93.5}\) & \(\bestgrp{95.9}\)                                                         & \(\bestgrp{92.5}\) & \(\bestgrp{92.4}\) & \(\bestgrp{93.0}\) \\
            XLM-R+SBERT                            & \(88.6\)           & \(90.8\)                                                                   & \(87.1\)           & \(88.6\)           & \(87.8\)           \\
            \bottomrule
        \end{tabular}
    }
    \caption{\label{tab:bucc}
        F1 scores on the BUCC bitext mining text.
        Simple average of scores from 4 tasks reported in the second column.
        Highest scores in their groups are {bolded}.
    }
\end{table}

Four models,
two unsupervised and two one-pair supervised
trained on either of the two language pairs,
are evaluated on two bitext mining or retrieval tasks
of the BUCC corpus (\cite{zweigenbaum2018multilingual}) and of the Tatoeba corpus (\cite{artetxe2019margin}).

\subsection{Baselines and comparisons}
\paragraph{Unsupervised baselines}
The XLM-R (\cite{conneau2020unsupervised}) bag-of-embedding (boe) representations out-of-the-box serve as the unsupervised baseline.
We identify the best-performing among the layers of orders of multiples of 4,
or layer \(L\in\lbrace 0, 4, 8, 12, 16, 20, 24 \rbrace\),
as the baseline.
In the case of BUCC mining task, for example, the best-performing baseline model is of layer 16 and denoted by XLM-R L16-boe.

Results from \textcite{kiros2020contextual}, \textcite{keung2020unsupervised}, and \textcite{kvapilikova2020unsupervised},
as state-of-the-art models for unsupervised bitext mining from pretrained language models,
are included for comparison
(see Section~\ref{paragraph:related work unsupervised parallel sentence mining} for a description of them).

\paragraph{Fully supervised models}
LASER (\cite{artetxe2019massively}) and LaBSE (\cite{feng2020languageagnostic}),
both fully supervised with multilingual bitexts,
are included for comparisons.
LASER is an LSTM-based encoder and translation model trained on parallel corpora of 93 languages,
and is the earlier leading system on the two mining tasks.
LaBSE on the other hand is a transformer-based multilingual sentence encoder supervised with parallel sentences from 109 languages
using the additive margin softmax (\cite{wang2018additive}) for the translation language modeling objective,
and has state-of-the-art performance on the two mining tasks.
Finally, XLM-R+SBERT from \cite{reimers2020making} is XLM-R fine-tuned to align representations of bitexts of 50 language pairs and to distill knowledge from
SBERT (\cite{reimers2019sentence}) fine-tuned on English paraphrase pairs.

\subsection{BUCC}

The {BUCC} corpora (\cite{zweigenbaum2018multilingual}),
consist of 95k to 460k sentences in each of 4 languages---German, French, Russian, and Mandarin Chinese---with around \(3\%\) of such sentences being English-aligned.
The task is to mine for the translation pairs.


\paragraph{Margin-based retrieval}
The retrieval is based on the margin-based similarity scores (\cite{artetxe2019margin}) related to CSLS (\cite{conneau2018word}),
\[
    \begin{aligned}
         & \func{score}(y^s, y^t) = \func{margin}(\func{sim}(y^s, y^t), \func{scale}(y^s, y^t)) \\
         & \func{scale}(y^s, y^t) =                                                             \\
         & \sum_{z \in \set{NN}_k(y^s)} \frac{\func{sim}(y^s, z)}{2k}
        + \sum_{z \in \set{NN}_k(y^t)} \frac{\func{sim}(y^t, z)}{2k}
        \, \text{,}
    \end{aligned}
\]
where \(\set{NN}_k (y)\) denotes the \(k\) nearest neighbors of \(y\) in the other language.
Here we use \(k=4\)
and the ratio margin function,
or \(\func{margin}(a, b) = a / b\),
following the literature (\cite{artetxe2019massively}).
By scaling up the similarity associated with more isolated embeddings,
margin-based retrieval helps alleviate the hubness problem (\cite{radovanovic2010hubs}),
where some embeddings or hubs are nearest neighbors of many other embeddings with high probability.

Following \textcite{hu2020xtreme},
our model is evaluated on the training split of the {BUCC} corpora,
and the threshold of the similarity score cutting off translations from non-translations is optimized for each language pair.
While \textcite{kvapilikova2020unsupervised} and \textcite{kiros2020contextual} optimize for the language-specific mining thresholds
as we do here,
\textcite{keung2020unsupervised} use a prior probability to infer the thresholds.
And different from all other baselines or models for comparisons presented here,
\textcite{kvapilikova2020unsupervised}'s model is evaluated upon the undisclosed test split of the BUCC corpus.

\paragraph{Results}

F1 scores on the BUCC dataset presented in Table~\ref{tab:bucc}
demonstrate that bilingual alignment learned by the model is transferable to other pairs of languages.
The hyperparameter values of the unsupervised model presented in the table are \(n=1, \alpha=0,  \lambda=5, \kappa=2\),
and those of the supervised model are \(n=1, \alpha=0\).

The adversarially-trained unsupervised model
outperforms the unsupervised baselines and nearing the state of the art,
and is thus effective in extracting sentence representations which are sharable across languages.
The choice of pivot pairs shows effects on the unsupervised models,
with the model trained on the de-en texts performing better than that on the ar-en texts
at mining for parallel sentences between English and German by 7 points.
The results suggest that
while alignment is transferable,
the unsupervised model can be further improved for multilingual alignment
by being trained on multilingual texts of more than two pivots.

The one-pair supervised model trained with bitexts of one pair of languages, on the other hand,
performs within a 6-point range of the fully supervised systems,
which shows that much alignment information from unsupervised pretrained models is recoverable by the simple extraction module.
Noticeably, the model supervised with ar-en bitexts but not from the four pairs of the task sees a 20-point increase from the plain XLM-R,
and the choice of dual pivots does not have significant effects on the supervised model.

\subsection{Tatoeba}

\begin{table}
    \centering
    {\small
        \setlength{\tabcolsep}{3pt} 
        \begin{tabular}{@{}llll@{}}
            \toprule
            Model                                  & \multicolumn{3}{c}{Average accuracy (\(\%\))}                                           \\
            \cmidrule{2-4}
                                                   & 36                                            & 41                 & 112                \\
            \midrule
            \emph{Unsupervised}                    &                                               &                    &                    \\
            Model (ar-en unsup.)                   & \(73.4\)                                      & \(70.8\)           & \(55.3\)           \\
            Model (ar-de unsup.)                   & \(71.5\)                                      & \(68.9\)           & \(53.3\)           \\
            Model (de-en unsup.)                   & \(\bestgrp{74.2}\)                            & \(\bestgrp{72.0}\) & \(\bestgrp{56.0}\) \\
                                                   &                                               &                    &                    \\
            XLM-R L12-boe                          & \(54.3\)                                      & \(53.1\)           & \(39.6\)           \\
            \textcite{kvapilikova2020unsupervised} & \(--\)                                        & \(50.2\)           & \(--\)             \\
            \midrule
            \emph{One-pair supervised}             &                                               &                    &                    \\
            Model (ar-en bitexts sup.)             & \(79.6\)                                      & \(78.3\)           & \(61.1\)           \\
            Model (ar-de bitexts sup.)             & \(74.1\)                                      & \(71.8\)           & \(56.0\)           \\
            Model (de-en bitexts sup.)             & \(\bestgrp{80.4}\)                            & \(\bestgrp{78.9}\) & \(\bestgrp{62.4}\) \\
            \midrule
            \emph{Supervised with 50+ pairs}       &                                               &                    &                    \\
            LASER                                  & \(85.4\)                                      & \(79.1\)           & \(66.9\)           \\
            LaBSE                                  & \(\bestgrp{95.0}\)                            & \(--\)             & \(\bestgrp{83.7}\) \\
            XLM-R+SBERT                            & \(86.2\)                                      & \(84.3\)           & \(67.1\)           \\
            \bottomrule
        \end{tabular}
    }
    \caption{\label{tab:tatoeba}
        Average accuracy scores on the Tatoeba dataset in three average groups.
        Highest scores in their groups are {bolded}.
    }
\end{table}

We also measure the parallel sentence matching accuracy over the Tatoeba dataset (\cite{artetxe2019massively}).
This dataset consists of 100 to 1,000 English-aligned sentence pairs for 112 languages,
and the task is to retrieve the translation in one of the target languages given a sentence in English
using absolute similarity scores without margin-scaling.

\paragraph{Results}

Matching accuracy for the retrieval task of the Tatoeba dataset are presented in Table~\ref{tab:tatoeba}.
Following \cite{feng2020languageagnostic},
average scores from different groups are presented to compare different models.
The 36 languages are those selected by Xtreme (\cite{hu2020xtreme}),
and the 41 languages are those for which results are presented in \cite{kvapilikova2020unsupervised}.
The hyperparameter values of the unsupervised model presented in the table are \(n=2, \alpha=0.2,  \lambda=10, \kappa=2\),
and those of the supervised model are \(n=1, \alpha=0\).

The unsupervised model outperforms the baselines by roughly 20 points,
and the one-pair supervised model performs close to the the supervised model LASER but falls shorts
by around 10 to 20 points to the other supervised model LaBSE.
When one of the two pivot languages is English, the choice of the pivot does not show much difference on this task on average.
While the models trained on ar-de (where neither pivot language is English) still exhibits strong transfer performance,
there is a drop of around 2 to 6 points from the models where English is one of the pivot languages (ar-en and de-en).

\section{Analysis}

To understand the factors affecting the performance of the model,
we consider several variants.
All models presented in this section are trained with the same hyperparameters presented in the evaluation section above
using either de-en corpora
or corpora of multiple language pairs (Section~\ref{sec:multipair}).

\subsection{Ablation}

\begin{table}
    \centering
    \begin{tabular}{@{}ll@{}}
        \toprule
        Model                                          & Tatoeba 36               \\
        \midrule
        \emph{Unsupervised}                            &                          \\
        XLM-R average-boe                              & \(54.9\)                 \\
        XLM-R L12-boe                                  & \(54.3\)                 \\
        \(\mathcal{L} = \mathcal{L}_\ind{adv}\)                                   \\
        \hspace{0.1in} linear combination              & \(34.7\)                 \\
        \hspace{0.1in} linear map                      & \(0.1\)                  \\
        \hspace{0.1in} linear combination + linear map & \(2.1\)                  \\
        \(\mathcal{L} = \mathcal{L}_\ind{cycle}\)                                 \\
        \hspace{0.1in} linear combination              & \(55.4\)                 \\
        \hspace{0.1in} linear map                      & \(69.8\)                 \\
        \hspace{0.1in} linear combination + linear map & \(68.0\)                 \\
        \(\mathcal{L} = \mathcal{L}_\ind{adv} + \lambda \mathcal{L}_\ind{cycle}\) \\
        \hspace{0.1in} linear combination              & \(47.4\)                 \\
        \hspace{0.1in} linear map                      & \(70.5\)                 \\
        \hspace{0.1in} linear combination + linear map & \(74.2\)                 \\
        \midrule
        \emph{One-pair supervised}                     &                          \\
        \(\mathcal{L} = \mathcal{L}_\ind{sup}\)                                   \\
        \hspace{0.1in} linear combination              & \(67.5\)                 \\
        \hspace{0.1in} linear map                      & \(80.1\)                 \\
        \hspace{0.1in} linear combination + linear map & \(80.4\)                 \\
        \bottomrule
    \end{tabular}
    \caption{\label{tab:ablation}
        Ablation results of the accuracy (\(\%\)) on the Tatoeba averaged across 36 languages.
    }
\end{table}

We ablate from the model the trainable components---the weighted linear combination and the linear map---as well as the two training losses, \(\mathcal{L}_\ind{adv}\) and \(\mathcal{L}_\ind{cycle}\),
of the unsupervised model.
When the weighted linear combination is ablated, we use
the unweighted average of embeddings across layers.

We evaluate on the average accuracy over the 36 languages of the Tatoeba corpus.
The results in Table~\ref{tab:ablation} show a few interesting trends.
First, the cycle consistency loss is essential for the unsupervised model, as can be seen by the very low performance when only $\mathcal{L}_\ind{adv}$ is used.
Secondly, the linear map plays a larger role than the linear combination in extracting alignment information in the unsupervised model:
in both conditions with cycle consistency loss, the linear map alone outperforms the linear combination alone,
and in the condition with only cycle consistency loss, the linear map alone does best.
Finally, in the one-pair supervised model,
the linear combination module alone shows gains of 13 points from the baseline
but does not produce gains when trained along with a linear map.

\subsection{Single-layer representations}

\begin{figure}
    \includegraphics[scale=0.62]{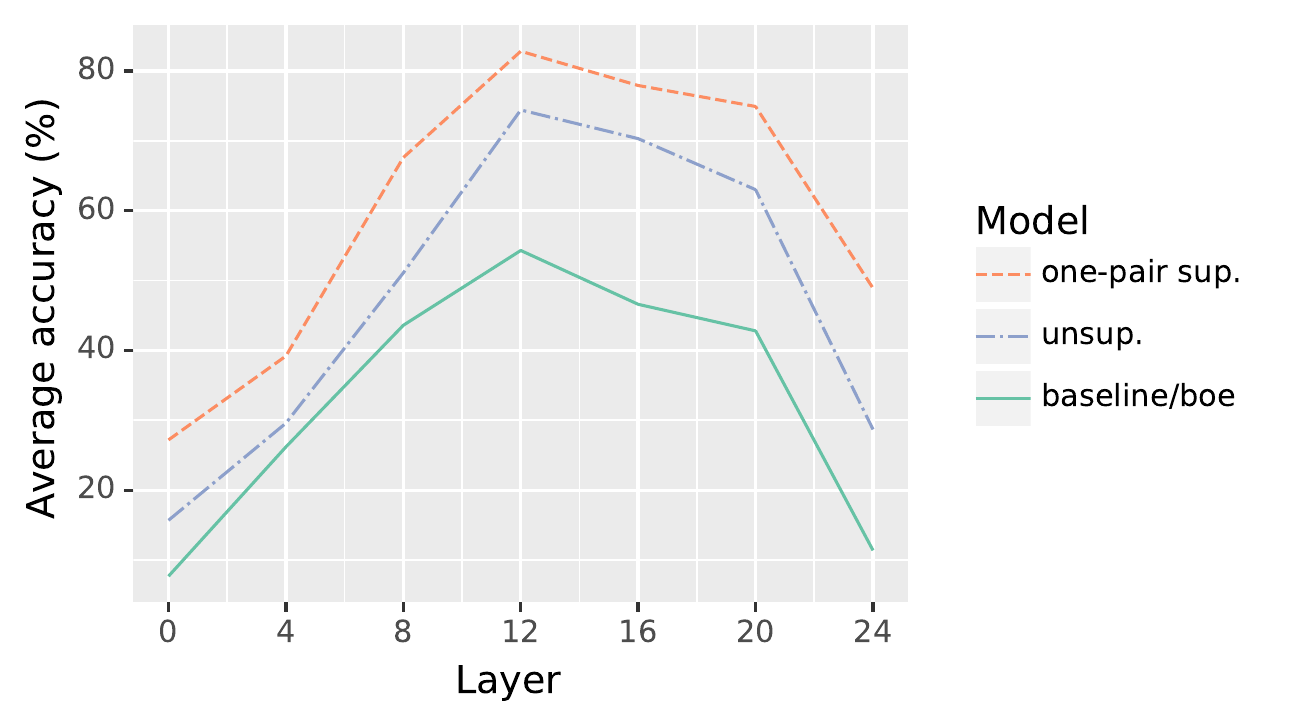}
    \caption{\label{fig:layers}
        Average accuracies of 36 languages of the Tatoeba task where the model is trained with representations from XLM-R layers of orders which are multiples of 4.
    }
\end{figure}

Previous studies show that representations from different layers differ in their cross-lingual alignment (\cite{pires2019multilingual, conneau2020emerging}).
To understand this phenomenon in the present setting,
we take layers whose orders are multiples of 4
and train the model with representations from one single layer without combining embeddings from different layers.

The average accuracy scores over 36 languages in Tatoeba summarized in Figure~\ref{fig:layers}
show that the alignment information is most extractable deep in the middle layers,
corroborating the findings from the previous work (\cite{kvapilikova2020unsupervised, litschko2021evaluating, chi2021xlme}).
The model trained with the best-performing layer shows similar or higher scores than the full model with learned linear combination,
which is consistent with the findings from the ablation that the learned linear combination is not essential for extracting alignment information.

\subsection{Training with multiple language pairs}
\label{sec:multipair}

\begin{table}
    \centering
    \begin{tabular}{@{}llllll@{}}
        \toprule
        Model        & \multicolumn{5}{c}{Model with \# of pairs }                                             \\
        \cmidrule{2-6}
                     & 1                                           & 2        & 4        & 8        & 16       \\
        \midrule
        Unsupervised & \(74.2\)                                    & \(74.9\) & \(75.2\) & \(77.5\) & \(77.1\) \\
        Supervised   & \(80.4\)                                    & \(80.0\) & \(80.0\) & \(83.3\) & \(85.7\) \\
        \bottomrule
    \end{tabular}
    \caption{\label{tab:multipair}
        Models with multiple pairs with accuracy (\(\%\)) on Tatoeba averaged across 36 languages.
    }
\end{table}

It is possible that a model trained on texts from more pairs of languages may improve upon the multilingual alignment so far demonstrated.
To test this,
we trained the model with the same hyperparameters as in the previous section but on texts from multiple languages,\footnote{The 16 pairs are those between \code{en} and \code{ar, bg, de, el, es, fa, fr, hi, id, ru, sw, th, tr, ur, vi, zh}.}
where each multi-pair model is trained on 16 million total sentences.
The results are in Table~\ref{tab:multipair}.
There is an aggregate 3 point improvement for the unsupervised model
and 5 point for the supervised model.
The results suggest that with our models,
one bilingual pivot is capable of extracting much transferable multilingual representations from XLM-R,
but using more pivots can still improve the transferability of the representations to some extent.

\subsection{Threshold transfer}

\begin{table}
    \centering
    \begin{tabular}{@{}lllll@{}}
        \toprule
        Model                      & \multicolumn{4}{c}{F1 score (\(\%\))\hspace*{1em} xx\(\leftrightarrow\)en}                                  \\
        \cmidrule{2-5}
                                   & de                                                                         & fr       & ru       & zh       \\
        \midrule
        \emph{Unsupervised}        &                                                                            &          &          &          \\
        Optimized thresholds       & \(91.4\)                                                                   & \(75.6\) & \(86.1\) & \(76.5\) \\
        Dual-pivot threshold       & \(91.4\)                                                                   & \(73.5\) & \(84.9\) & \(75.8\) \\
        \midrule
        \emph{One-pair supervised} &                                                                            &          &          &          \\
        Optimized thresholds       & \(92.5\)                                                                   & \(89.6\) & \(90.3\) & \(85.8\) \\
        Dual-pivot threshold       & \(92.5\)                                                                   & \(89.6\) & \(90.2\) & \(85.2\) \\
        \bottomrule
    \end{tabular}
    \caption{\label{tab:threshold}
        F1 scores on the BUCC training split with the model trained with de-en texts.
        The system with optimized thresholds tunes the threshold for each pair, as in Table~\ref{tab:bucc};
        the dual-pivot system uses the threshold from the de-en pair for all four pairs.
    }
\end{table}

Previous work observes that in the BUCC mining task, the thresholds optimized for different language pairs are close to one another,
suggesting that one can tune the threshold on high-resource pairs and use the system to mine other language pairs (\cite{kiros2020contextual}).
We examine the mining performance on the BUCC dataset of two threshold schemes: optimized thresholds, where thresholds are optimized for each language pair, and the dual-pivot threshold, where the threshold optimized for the pivot pair, in this case German-English, is used to mine all languages.

The scores from these two schemes on BUCC are summarized in Table~\ref{tab:threshold}.
The results show that the thresholds optimized for the pivot pair transfer to other pairs with at most a 2-point decrease in the F1 score,
and that the results of the two schemes are almost identical to the one-pair supervised model.
These experiments corroborate the previous observation
and demonstrate yet another case for leveraging texts from resource-rich pairs for unsupervised mining of other language pairs.

\subsection{Multilingual semantic textual similarity}

\begin{table}
    \setlength\tabcolsep{5pt}
    \centering
    {\small
        \begin{tabular}{@{}lllllll@{}}
            \toprule
            Model & ar-ar    & en-en    & es-es    & en-ar    & en-es    & en-tr    \\
            \midrule
            \emph{Unsup.}                                                           \\
            XLM-R & \(47.2\) & \(58.5\) & \(53.0\) & \(31.2\) & \(17.3\) & \(28.1\) \\
            Model & \(51.7\) & \(74.5\) & \(63.8\) & \(42.0\) & \(41.9\) & \(37.9\) \\ 
            \midrule
            \emph{One-pair sup.}                                                    \\
            Model & \(54.7\) & \(67.5\) & \(65.0\) & \(43.7\) & \(39.8\) & \(44.0\) \\
            \midrule
            \emph{Other systems}                                                    \\
            LASER & \(68.9\) & \(77.6\) & \(79.7\) & \(66.5\) & \(57.9\) & \(72.0\) \\
            LaBSE & \(69.1\) & \(79.4\) & \(80.8\) & \(74.5\) & \(65.5\) & \(72.0\) \\
            SBERT & \(79.6\) & \(88.8\) & \(86.3\) & \(82.3\) & \(83.1\) & \(80.9\) \\
            \bottomrule
        \end{tabular}
    }
    \caption{\label{tab:sts}
        Spearman correlation (\(\%\)) on multilingual STS 2017 (\cite{cer2017semeval}) of the unsupervised and supervised models trained on the de-en pair.
    }
\end{table}

To test whether our method of training language-agnostic sentence encoders encourages meaning-based representations,
we evaluate the models on the multilingual semantic textual similarity (STS) 2017 (\cite{cer2017semeval}) with Spearman correlation reported in Table~\ref{tab:sts}. All evaluation pairs on average see about \(10\) percentage-point increases from baseline (XLM-R L12-boe) for both models.
Yet the gaps between our models and the fully supervised systems suggest
that supervision with more language pairs and more trainable parameters likely encourages sentence representations to be closer to what humans see as meaning.

\section{Conclusion}
This work shows that training for bilingual alignment benefits multilingual alignment for unsupervised bitext mining.
The unsupervised model shows the effectiveness of adversarial training with cycle consistency for building multilingual language models,
and reaches the state of the art of unsupervised bitext mining.
Both unsupervised and one-pair supervised models show
that significant multilingual alignment in an unsupervised language model can be recovered by a linear mapping,
and that combining monolingual and bilingual training data can be a data-efficient method for promoting multilingual alignment.
Future work may combine both the supervised and the unsupervised techniques to attain sentence embeddings with stronger multilingual alignment
through the transferability of bilingual alignment demonstrated in this work,
and such work will benefit tasks involving languages of low resources in bitexts.

\bibliography{bib/bibliography}


\end{document}